\documentclass[11pt]{article}

\usepackage[T1]{fontenc}
\usepackage{lmodern}        
\usepackage[utf8]{inputenc} 

\usepackage[english]{babel}

\usepackage[letterpaper,top=2cm,bottom=2cm,left=3cm,right=3cm,marginparwidth=1.75cm]{geometry}

\usepackage{amsmath,amssymb,amsthm}
\usepackage{graphicx}
\usepackage{subcaption}
\usepackage{float}

\usepackage[colorlinks=true,
            linkcolor=blue,
            citecolor=blue,
            urlcolor=blue]{hyperref}

\usepackage[numbers,sort&compress]{natbib}

\usepackage{booktabs}
\usepackage{array}
\usepackage{caption}
\usepackage{subcaption}
\usepackage{url}

\usepackage{algorithm}
\usepackage{algpseudocode}
\usepackage{multirow}

\title{Exploring the Hierarchical Reasoning Model for Small Natural-Image Classification Without Augmentation}
\author{Alexander V. Mantzaris}

\date{%
  Department of Data, Mathematical and Statistical Sciences, University of Central Florida, USA\\[1ex]%
  \today
}

\begin{document}
\maketitle

\begin{abstract}
This paper asks whether the Hierarchical Reasoning Model (HRM) with the two Transformer-style modules $(f_L,f_H)$, one step (DEQ-style) training, deep supervision, Rotary Position Embeddings, and RMSNorm can serve as a practical image classifier. It is evaluated on MNIST, CIFAR-10, and CIFAR-100 under a deliberately raw regime: no data augmentation, identical optimizer family with one-epoch warmup then cosine-floor decay, and label smoothing. HRM optimizes stably and performs well on MNIST ($\approx 98\%$ test accuracy), but on small natural images it overfits and generalizes poorly: on CIFAR-10, HRM reaches 65.0\% after 25 epochs, whereas a two-stage Conv--BN--ReLU baseline attains 77.2\% while training $\sim 30\times$ faster per epoch; on CIFAR-100, HRM achieves only 29.7\% test accuracy despite 91.5\% train accuracy, while the same CNN reaches 45.3\% test with 50.5\% train accuracy. Loss traces and error analyses indicate healthy optimization but insufficient image-specific inductive bias for HRM in this regime. It is concluded that, for small-resolution image classification without augmentation, HRM is not competitive with even simple convolutional architectures as the HRM currently exist but this does not exclude possibilities that modifications to the model may allow it to improve greatly.
\end{abstract}

\section{Introduction}

Reasoning remains a central challenge in machine learning. A dominant line of work elicits reasoning in large language models (LLMs) via chain-of-thought (CoT) prompting, which provides explicit intermediate steps at inference time~\cite{wei2022cot}. In contrast, the recently introduced \emph{Hierarchical Reasoning Model} (HRM) proposes a compact, recurrent architecture that learns to carry out multi-step reasoning \emph{within a single forward pass}, without external CoT supervision~\cite{wang2025hrm}. HRM consists of two interdependent modules operating at different time scales: a high-level planner (\(f_H\)) and a low-level executor (\(f_L\)), which exchange state over several cycles. Training leverages a one-step gradient derived from deep equilibrium models (DEQs)---using implicit differentiation to bypass full backpropagation through time---thereby achieving constant memory with respect to effective depth~\cite{bai2019deq,wang2025hrm}. In practice, HRM employs modern Transformer components (eg., RMSNorm and rotary position embeddings) to stabilize and structure the dynamics~\cite{zhang2019rmsnorm,su2021roformer}.

The initial HRM report demonstrated striking sample efficiency and performance on algorithmic and visual reasoning benchmarks (eg, Sudoku, maze planning, and the Abstraction and Reasoning Corpus, ARC) using comparatively small models and limited data, spurring rapid community interest and open-source reimplementations~\cite{wang2025hrm,sapient2025hrmcode,chollet2019measure}. Beyond its empirical results, HRM is conceptually appealing: it offers a brain-inspired, multi-timescale recurrent formulation with a mathematically grounded training rule (via DEQ), hinting at a path toward controllable computation depth and amortized planning inside the network. Given this novelty and promise, it is valuable to probe HRM’s \emph{inductive biases} outside its original evaluation settings. In this work the study of the HRM in small-resolution image classification (MNIST \cite{lecun1998gradient,lecun2010mnist}, CIFAR-10/100\cite{krizhevsky2009learning}) under a deliberately raw straightforward training regime (no augmentation) and compare it against a conventional convolutional baseline. The goal is to clarify where HRM’s architectural design helps or hinders generalization. (code is available at \url{github.com/mantzaris/ImagesHRM})

\section{Methodology}

The HRM can be described abstractly by coupled updates between a high–level state and a low–level state, with a separation of time scales:
\begin{equation}
\mathbf{z}^{t+1}_{L} = f_{L}\!\left(\mathbf{z}^{t}_{L}, \mathbf{z}^{t}_{H}, \tilde{\mathbf{x}};\, \theta_{L}\right),
\qquad
\mathbf{z}^{t+1}_{H} = f_{H}\!\left(\mathbf{z}^{t}_{H}, \mathbf{z}^{t+1}_{L};\, \theta_{H}\right).
\end{equation}
These equations express a single alternating update at step \(t\), where the low–level state is updated first from its previous value, the current high–level state, and the tokenized input, followed by an update of the high–level state conditioned on the newly refined low–level state. In the image specialization above, the same principle is implemented with \(T\) low–level micro–updates per high–level macro–update, producing the cycle and segment schedules used in training.

The adaptation for images thus consists of:  patch–tokenization with a \([\mathrm{CLS}]\) channel. Then two Transformer modules \(f_{L}\) and \(f_{H}\) operating at different time scales with element–wise input fusion. Then a one–step gradient approximation that backpropagates only through the last low– and high–level updates per segment, and the deep supervision over segments with detached state carryover. The classifier head reads the high–level \([\mathrm{CLS}]\) representation to produce logits for cross–entropy training.

Hierarchical Reasoning Model (HRM) adapted for image classification consists of two interdependent recurrent modules operating at distinct time scales: a low–level perceptual module \(f_{L}\) and a high–level reasoning module \(f_{H}\). Both \(f_{L}\) and \(f_{H}\) are instantiated as Transformer encoders with identical dimensionality, bias–free linear layers, RMS normalization, rotary position embedding within multi–head self–attention, and GEGLU feed–forward sublayers. Inputs to each module are merged by element–wise addition before the Transformer stack, and the output head \(f_{O}\) produces class logits from the high–level state.

\begin{equation}
\mathbf{x} \in \mathbb{R}^{B \times H_{\text{img}} \times W_{\text{img}} \times C_{\text{in}}}, 
\qquad 
y \in \{0,\dots, K-1\}.
\end{equation}
The dataset consists of batches of images \(\mathbf{x}\) and labels \(y\). The goal is to learn parameters \(\theta = \{\theta_{I}, \theta_{L}, \theta_{H}, \theta_{O}\}\) that minimize the supervised loss on \(K\) classes.

\begin{equation}
\tilde{\mathbf{x}} = f_{I}(\mathbf{x}; \theta_{I}) \in \mathbb{R}^{B \times S \times D}.
\end{equation}
Images are tokenized by \(f_{I}\) into a sequence of \(S\) tokens of width \(D\). Using non–overlapping patch embedding with a prepended \([\mathrm{CLS}]\) token so that \(S = 1 + (H_{\text{img}}/P)(W_{\text{img}}/P)\). This step converts spatial inputs into a sequence representation suitable for Transformer processing.

\begin{equation}
\mathbf{z}^{0}_{H} \sim \mathcal{TN}(0,1; -2, 2), 
\qquad 
\mathbf{z}^{0}_{L} \sim \mathcal{TN}(0,1; -2, 2),
\qquad 
\mathbf{z}^{0}_{H}, \mathbf{z}^{0}_{L} \in \mathbb{R}^{B \times S \times D}.
\end{equation}
The initial high–level and low–level states are sampled once from a truncated normal distribution and kept fixed through training. This matches the HRM practice of using fixed random initial states for recurrent inference.

\begin{equation}
\mathbf{z}^{i,\tau}_{L} 
= 
f_{L}\!\left( \mathbf{z}^{i,\tau-1}_{L} + \mathbf{z}^{i-1}_{H} + \tilde{\mathbf{x}} ;\, \theta_{L} \right),
\quad 
\tau = 1,\dots,T,
\quad 
i = 1,\dots,N.
\end{equation}
Within the \(i\)-th high–level cycle, the low–level module performs \(T\) fast updates. Each update fuses the previous low–level state, the last available high–level state, and the tokenized image via element–wise addition before the Transformer encoder \(f_{L}\). This captures iterative perceptual refinement conditioned on the current global context.

\begin{equation}
\mathbf{z}^{i}_{H} 
= 
f_{H}\!\left( \mathbf{z}^{i-1}_{H} + \mathbf{z}^{i,T}_{L} ;\, \theta_{H} \right),
\quad 
i = 1,\dots,N.
\end{equation}
At the end of each cycle, the high–level module updates once from its previous state and the final low–level state of the cycle. This realizes hierarchical convergence, where slow global reasoning is informed by fast perceptual updates.

\begin{equation}
\boldsymbol{\ell} 
= 
f_{O}\!\left( \mathbf{z}^{N}_{H}[:,0,:] ;\, \theta_{O} \right) 
\in \mathbb{R}^{B \times K},
\qquad 
\hat{y} = \arg\max_{k} \boldsymbol{\ell}_{:,k}.
\end{equation}
The output head \(f_{O}\) maps the high–level \([\mathrm{CLS}]\) channel to logits over classes. The prediction \(\hat{y}\) is the index of the maximum logit per example.

\begin{equation}
\mathcal{L}_{\text{CE}} 
= 
-\frac{1}{B} \sum_{b=1}^{B} \log 
\frac{\exp\big(\boldsymbol{\ell}_{b,\,y_{b}}\big)}{\sum_{k=1}^{K} \exp\big(\boldsymbol{\ell}_{b,\,k}\big)}.
\end{equation}
The training objective is the average cross–entropy between predicted logits and ground truth labels. This objective is computed at the end of each supervised segment, described below.

\begin{equation}
\left( \mathbf{z}^{\ast}_{L}, \mathbf{z}^{\ast}_{H} \right) 
= 
\Phi\!\left( \tilde{\mathbf{x}}, \mathbf{z}^{0}_{L}, \mathbf{z}^{0}_{H};\, \theta \right),
\quad
\frac{\partial \mathcal{L}_{\text{CE}}}{\partial \theta}
\approx
\frac{\partial \mathcal{L}_{\text{CE}}}{\partial \left( \mathbf{z}^{\ast}_{L}, \mathbf{z}^{\ast}_{H} \right)}
\cdot
\frac{\partial \left( \mathbf{z}^{\ast}_{L}, \mathbf{z}^{\ast}_{H} \right)}{\partial \theta}\Big|_{\text{last step}}.
\end{equation}
Training uses a one–step (DEQ–style) gradient approximation. All intermediate low–level and high–level updates within a segment are treated as constants for backpropagation, and gradients are taken only through the final low–level update and the final high–level update of the segment. This avoids backpropagation through many recurrent steps while preserving a direct gradient path from the loss to the parameters at the end of the segment.

\begin{equation}
\mathcal{L}_{\text{segment}}
=
\mathcal{L}_{\text{CE}}\!\left( f_{O}\!\left( \mathbf{z}^{N}_{H}[:,0,:] ; \theta_{O} \right),\, y \right),
\qquad
\mathcal{L}_{\text{deep}} 
= 
\sum_{m=1}^{M} \lambda_{m}\, \mathcal{L}^{(m)}_{\text{segment}}.
\end{equation}
Deep supervision is implemented by splitting the forward process into \(M\) segments per batch. After each segment, a supervised loss is computed, an optimizer step is taken, and the recurrent state is detached before continuing to the next segment. The total training objective is a weighted sum of the segment losses.

\begin{equation}
\mathbf{q} 
= 
\sigma\!\left( f_{Q}\!\left( \mathbf{z}^{N}_{H}[:,0,:] ; \theta_{Q} \right) \right) 
\in \mathbb{R}^{B \times 2},
\qquad
\pi(\text{halt}\mid \mathbf{z}^{N}_{H}) = \mathbf{q}_{:,0},
\quad
\pi(\text{continue}\mid \mathbf{z}^{N}_{H}) = \mathbf{q}_{:,1}.
\end{equation}
An optional halting head \(f_{Q}\) can be trained to predict whether to halt or to continue segments adaptively (Adaptive Computation Time). In these basic experiments this head is disabled during training and can be used at inference to cap or adapt the number of segments.
\begin{equation}
f_{L}(\cdot;\theta_{L}) = \text{TransformerEncoder}(D,\ H,\ L_{L}).
\end{equation}
\begin{equation}
f_{H}(\cdot;\theta_{H}) = \text{TransformerEncoder}(D,\ H,\ L_{H}).
\end{equation}
Each \(\text{TransformerEncoder}\) uses identical dimensionality \(D\) and number of heads \(H\), with depth \(L_{L}\) for \(f_{L}\) and \(L_{H}\) for \(f_{H}\). A single encoder block applies, in order, RMS normalization, multi-head self-attention with rotary position embedding, RMS normalization, and a GEGLU feed-forward layer.
Both modules are encoder–only Transformer stacks with the same model width \(D\) and number of heads, and they differ only in their parameters \(\theta_{L}\) and \(\theta_{H}\). Inputs to each stack are summed element–wise before entering the first attention layer.

\section{Results}
Here the results of the training and testing are presented for the HRM on images.

\subsection{MNIST}

This experiment applies the faithful HRM to the MNIST digit classification task. Both the low–level module \(f_{L}\) and the high–level module \(f_{H}\) are instantiated as Transformer encoders with identical model width and number of heads. Training follows the one–step (DEQ–style) gradient with deep supervision over segments. Images are tokenized into non–overlapping patches with a prepended \([\mathrm{CLS}]\) token; the classifier head \(f_{O}\) reads the \([\mathrm{CLS}]\) channel of the final high–level state to produce logits.

\paragraph{Results.}
Training for three epochs yields a final test accuracy of \(0.9801\) with \(1{,}053{,}056\) parameters. The per–epoch training metrics are summarized in Table~\ref{tab:mnist-hrm-metrics}.

\begin{table}[tb]
\centering
\caption{MNIST training metrics for the HRM.}
\label{tab:mnist-hrm-metrics}
\begin{tabular}{lccc}
\toprule
Epoch & Train loss & Train accuracy & Test accuracy (final) \\
\midrule
01 & 0.3969 & 0.8695 & \multirow{3}{*}{0.9801} \\
02 & 0.0739 & 0.9778 & \\
03 & 0.0579 & 0.9828 & \\
\bottomrule
\end{tabular}
\end{table}

\begin{figure}[tb]
  \centering
  \includegraphics[width=0.9\linewidth]{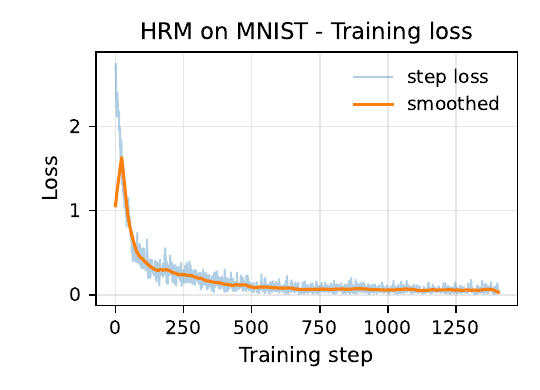}
  \caption{Training loss across optimization steps. The lightly shaded trace is the instantaneous per–segment step loss; the darker trace is a moving average for readability. Loss stabilizes after the initial descent, aligning with the steady gains in accuracy.}
  \label{fig:mnist-loss}
\end{figure}

\begin{figure}[tb]
  \centering
  \includegraphics[width=0.9\linewidth]{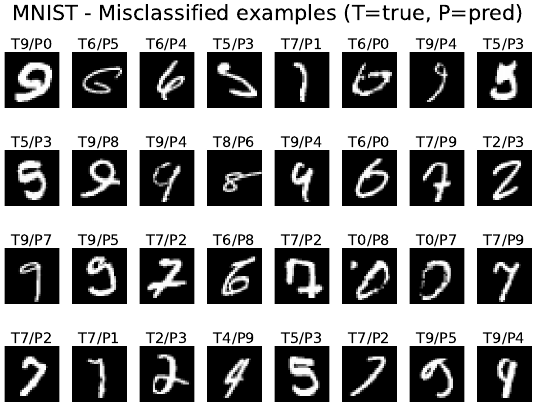}
  \caption{Misclassified MNIST examples. Each tile shows the true class \(T\) and the predicted class \(P\). Confusions concentrate on digit pairs with similar topology (for example, curled tails in \(9\) vs.\ closed loops in \(0\); angled \(7\) vs.\ vertical \(1\)).}
  \label{fig:mnist-errors}
\end{figure}

Figure~\ref{fig:mnist-loss} shows the loss curve across optimization steps. A rapid decrease occurs during the first epoch, followed by a smooth plateau, consistent with the epoch–level improvements in Table~\ref{tab:mnist-hrm-metrics}. Figure~\ref{fig:mnist-errors} displays representative misclassifications; most errors occur between visually similar digits (for example, \(9\) vs.\ \(0\), \(7\) vs.\ \(1\), \(5\) vs.\ \(3\)), indicating residual pixel–level ambiguity rather than a systematic modeling failure.

Overall, the quantitative trajectory (loss and accuracy per epoch) and the qualitative error patterns are consistent with an HRM that alternates fast low–level perceptual updates with slower high–level consolidation. The one–step gradient with deep supervision provides stable optimization without backpropagation through the entire reasoning horizon, achieving competitive performance on MNIST with a small parameter budget.

\subsection{CIFAR-10}

The evaluation of the HRM on CIFAR-10 is done using the image classification setting without data augmentation.
Inputs are per-channel standardized RGB images of shape $32{\times}32{\times}3$.
Tokenization uses non-overlapping $4{\times}4$ patches, yielding a sequence length of $1{+}8{\times}8{=}65$ tokens (including a learned \texttt{[CLS]} token).
Both the low- and high-level modules ($f_L,f_H$) are encoder-only Transformer stacks with shared dimensionality: $d_{\mathrm{model}}{=}192$, $n_{\mathrm{heads}}{=}6$ (per-head dimension $32$), MLP multiplier $4$, and $3$ layers in each of $f_L$ and $f_H$.
Training the HRM recurrent schedule of $N{=}2$ high-level cycles and $T{=}3$ low-level updates per cycle, deep supervision with $M_{\mathrm{train}}{=}2$ segments (state detached between segments), and the one-step gradient approximation (all but the final $L/H$ updates run under \texttt{stop\_gradient}).
At evaluation running $M_{\mathrm{eval}}{=}3$ segments with fresh initial states per batch.
Optimization uses AdamW (global-norm clipping $1.0$), linear warmup for one epoch followed by cosine decay with a $20\%$ floor; learning rate $3{\times}10^{-4}$, weight decay $5{\times}10^{-4}$, batch size $128$, label smoothing $\varepsilon{=}0.05$, and no dropout or stochastic depth.
Training runs for $25$ epochs with the above settings and no adaptive computation/halting.

\begin{figure}[tb]
  \centering
  \includegraphics[width=\linewidth]{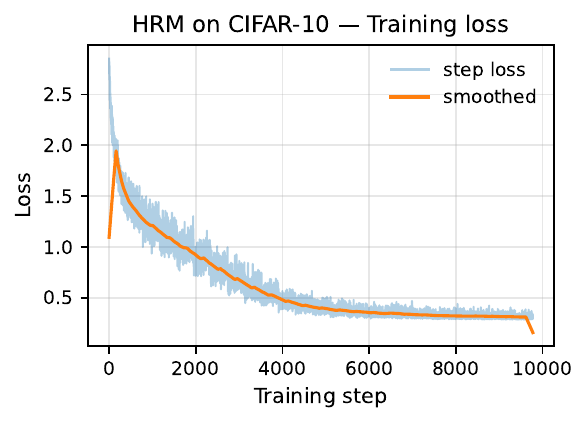}
  \caption{HRM on CIFAR-10: training loss per step (light) with a moving-average smoothed trace (dark). The curve decreases smoothly without instabilities.}
  \label{fig:hrm-cifar10-loss}
\end{figure}

\begin{figure}[tb]
  \centering
  \includegraphics[width=\linewidth]{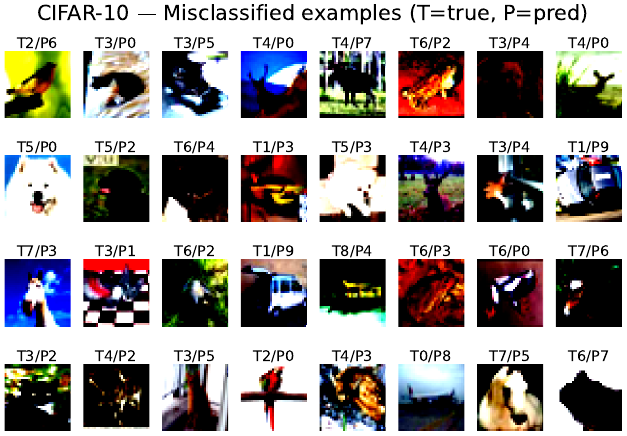}
  \caption{HRM on CIFAR-10: examples misclassified by the final model. Each tile shows the ground-truth (\texttt{T}) and prediction (\texttt{P}) indices.}
  \label{fig:hrm-cifar10-errors}
\end{figure}

The loss curve in Fig.~\ref{fig:hrm-cifar10-loss} shows a steady, monotonic decrease, indicating stable optimization under the one-step gradient with deep supervision. Nevertheless, the model quickly overfits: by epochs 21--25 the training accuracy saturates at $\approx 0.99$ while generalization remains poor (see also the diverse confusions visible in the misclassified grid in Fig.~\ref{fig:hrm-cifar10-errors}).\footnote{Typical epoch throughput in the runs was $\sim$391 steps/epoch at $\sim$1.11 it/s, i.e., $\approx$5:53 per epoch.}
The combination of high training accuracy and low test accuracy indicates that, in this raw (no-augmentation) regime on small natural images, HRM acts as a high-capacity learner without a strong image-specific inductive bias.

In this configuration HRM overfits rapidly and, despite lengthy training, achieves only $65.04\%$ test accuracy after $25$ epochs, consistent with the qualitative errors in Fig.~\ref{fig:hrm-cifar10-errors}. Taken together with the smooth loss in Fig.~\ref{fig:hrm-cifar10-loss}, this suggests optimization is stable but the model lacks the inductive bias needed for CIFAR-10 in the no-augmentation regime. The training accuracy was maintained to be close too 100\% throughout the later epochs.

\subsubsection{CIFAR-10: CNN baseline}

Using a classic small convolutional network: two Conv-BN-ReLU blocks with $3{\times}3$ kernels per stage followed by $2{\times}2$ max pooling (spatial resolution $32{\times}32 \rightarrow 16{\times}16 \rightarrow 8{\times}8$), then global average pooling and a linear classifier. The base width is $64$ channels in the first stage and $128$ in the second; there is no dropout. Inputs are per-channel standardized RGB images ($32{\times}32{\times}3$) with \emph{no data augmentation}. Optimization mirrors the HRM runs: AdamW with linear warmup (one epoch) followed by cosine decay with a $20\%$ floor, global-norm clipping $1.0$, label smoothing $\varepsilon=0.05$, learning rate $3{\times}10^{-4}$, weight decay $5{\times}10^{-4}$, batch size $128$, and $25$ training epochs. The final model has $261{,}824$ parameters. Typical epoch throughput is $\sim$391 steps at $\sim$31 it/s (about $12$ s/epoch).

\begin{figure}[tb]
  \centering
  \includegraphics[width=\linewidth]{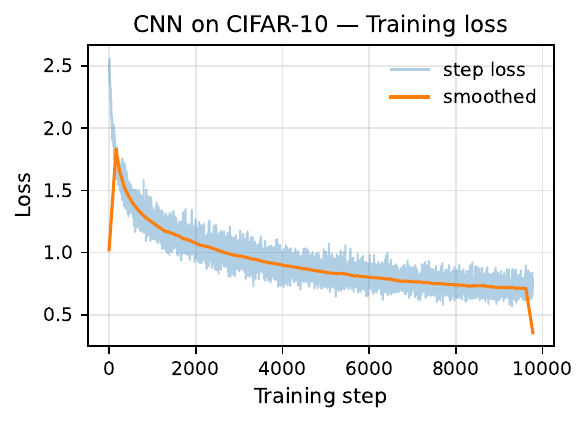}
  \caption{CNN on CIFAR-10: training loss per step (light) with a moving-average smoothed trace.}
  \label{fig:cnn-cifar10-loss}
\end{figure}

\begin{figure}[tb]
  \centering
  \includegraphics[width=\linewidth]{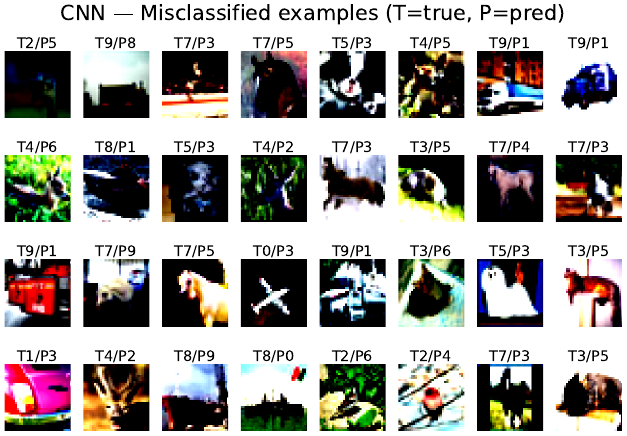}
  \caption{CNN on CIFAR-10: examples misclassified by the final model. Each tile shows ground-truth (\texttt{T}) and prediction (\texttt{P}).}
  \label{fig:cnn-cifar10-errors}
\end{figure}

Table~\ref{tab:cifar10-cnn-vs-hrm} aggregates the end-of-training epoch metrics for both models. HRM saturates to near-perfect training accuracy by epoch~25 but generalizes poorly; the CNN improves steadily and achieves a substantially higher test accuracy with far shorter epoch wall-time.
\begin{table}[t]
  \centering
  \begin{tabular}{@{}rcccccc@{}}
    \toprule
    & \multicolumn{3}{c}{\textbf{HRM}} & \multicolumn{3}{c}{\textbf{CNN}} \\
    \cmidrule(lr){2-4} \cmidrule(lr){5-7}
    \textbf{Epoch} & \textbf{Train loss} & \textbf{Train acc} & \textbf{Test acc} &
                      \textbf{Train loss} & \textbf{Train acc} & \textbf{Test acc} \\
    \midrule
    18 & 0.3383 & 0.9801 & ---    & 0.7699 & 0.8311 & 0.7452 \\
    19 & 0.3317 & 0.9833 & ---    & 0.7601 & 0.8362 & 0.7553 \\
    20 & 0.3252 & 0.9850 & ---    & 0.7506 & 0.8399 & 0.7604 \\
    21 & 0.3218 & 0.9865 & ---    & 0.7400 & 0.8434 & 0.7701 \\
    22 & 0.3198 & 0.9875 & ---    & 0.7325 & 0.8470 & 0.7726 \\
    23 & 0.3175 & 0.9882 & ---    & 0.7252 & 0.8518 & 0.7644 \\
    24 & 0.3141 & 0.9892 & ---    & 0.7195 & 0.8538 & 0.7744 \\
    25 & 0.3117 & 0.9900 & \textbf{0.6504} & 0.7125 & 0.8586 & \textbf{0.7722} \\
    \bottomrule
  \end{tabular}
  \caption{CIFAR-10 (no augmentation). Late-epoch metrics for HRM and the CNN baseline under identical optimization settings and preprocessing. HRM overfits rapidly (near-1.0 train accuracy) yet attains only $65.04\%$ test accuracy at epoch 25; the CNN reaches $77.22\%$ while training much faster per epoch.}
  \label{tab:cifar10-cnn-vs-hrm}
\end{table}

Under the same regime (no augmentation, identical optimizer family, label smoothing), this CNN baseline attains \textbf{77.22\%} test accuracy and trains roughly \textbf{$\sim$30$\times$ faster per epoch} than HRM, whereas HRM reaches only \textbf{65.04\%} after $25$ epochs despite near-saturated training accuracy ($\approx 0.99$ over epochs 21--25).%
As shown by the HRM loss curve and error grid (Figs.~\ref{fig:hrm-cifar10-loss} and~\ref{fig:hrm-cifar10-errors} in the previous subsection), HRM optimizes smoothly but overfits quickly, while the CNN exhibits better generalization on CIFAR-10 in this setting (Figs.~\ref{fig:cnn-cifar10-loss} and~\ref{fig:cnn-cifar10-errors}).%
Overall, these results indicate that \emph{with no augmentation and minimal regularization on small natural images}, a simple convolutional architecture is substantially more suitable than HRM for CIFAR-10 classification.

\subsection{CIFAR-100}

The evaluation of the HRM on CIFAR-100 follows the same no-augmentation image-classification protocol as CIFAR-10.
Inputs are per-channel standardized RGB images of shape $32{\times}32{\times}3$.
Tokenization uses non-overlapping $4{\times}4$ patches, yielding a sequence length of $1{+}8{\times}8{=}65$ tokens (including a learned \texttt{[CLS]} token).
Both the low- and high-level modules ($f_L,f_H$) are encoder-only Transformer stacks with shared dimensionality: $d_{\mathrm{model}}{=}192$, $n_{\mathrm{heads}}{=}6$ (per-head dimension $32$), MLP multiplier $4$, and $3$ layers in each of $f_L$ and $f_H$.
Training uses the HRM recurrent schedule of $N{=}2$ high-level cycles and $T{=}3$ low-level updates per cycle, deep supervision with $M_{\mathrm{train}}{=}2$ segments (state detached between segments), and the one-step gradient approximation (all but the final $L/H$ updates run under \texttt{stop\_gradient}).
At evaluation time, $M_{\mathrm{eval}}{=}3$ segments are run with fresh initial states per batch.
Optimization is AdamW (global-norm clipping $1.0$), linear warmup for one epoch followed by cosine decay with a $20\%$ floor; learning rate $3{\times}10^{-4}$, weight decay $5{\times}10^{-4}$, batch size $128$, label smoothing $\varepsilon{=}0.05$, and no dropout or stochastic depth.
Training is run for $25$ epochs with the above settings and no adaptive computation/halting.

\begin{figure}[tb]
  \centering
  \includegraphics[width=\linewidth]{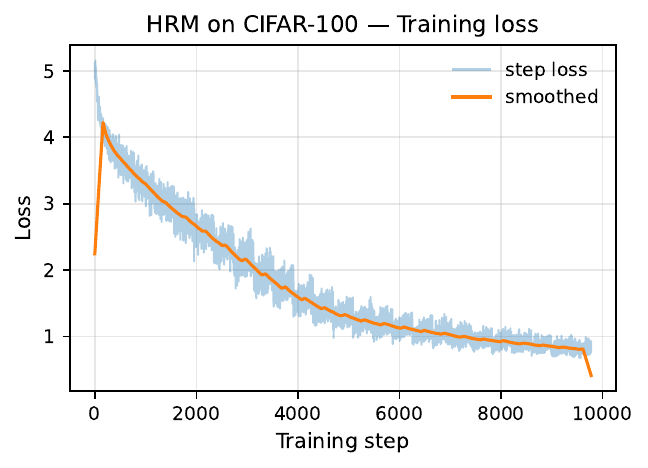}
  \caption{HRM on CIFAR-100: training loss per step (light) with a moving-average smoothed trace.
  The curve decreases smoothly without instabilities, indicating stable optimization.}
  \label{fig:hrm-cifar100-loss}
\end{figure}

\begin{figure}[tb]
  \centering
  \includegraphics[width=\linewidth]{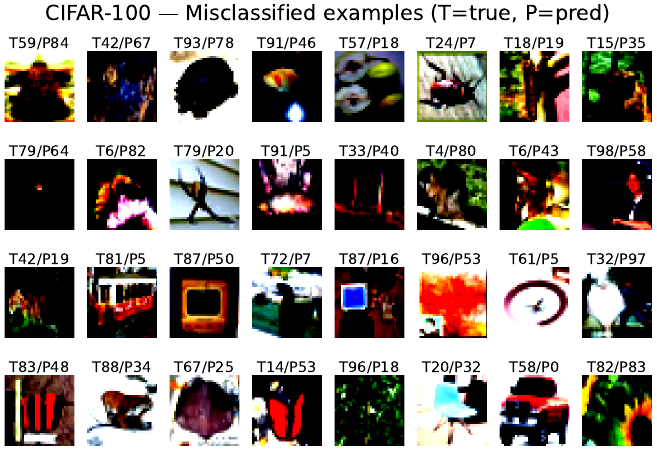}
  \caption{HRM on CIFAR-100: examples misclassified by the final model.
  Each tile shows the ground-truth (\texttt{T}) and prediction (\texttt{P}) indices.}
  \label{fig:hrm-cifar100-errors}
\end{figure}

The loss trace in Fig.~\ref{fig:hrm-cifar100-loss} shows a steady, monotonic decrease across training, similar to the CIFAR-10 behavior, which confirms that the one-step gradient with deep supervision optimizes stably in this setting.
However, the model rapidly overfits on CIFAR-100: by epoch~25, the training accuracy reaches $91.53\%$ while the test accuracy is only \textbf{29.70\%}.
This widening train--test gap, together with the diverse misclassifications in Fig.~\ref{fig:hrm-cifar100-errors}, indicates that in the no-augmentation regime on a $100$-class natural-image task, HRM functions as a high-capacity learner without the inductive bias needed for good generalization.
Typical epoch throughput was $\sim$391 steps/epoch at $\sim$1.11 it/s (about $5{:}52$ per epoch), so extending training beyond $25$ epochs would primarily increase training accuracy and wall-clock time without addressing the lack of generalization.

\subsubsection{CIFAR-100: CNN baseline}

Using the same small convolutional network as in the CIFAR-10 baseline: two Conv--BN--ReLU blocks with $3{\times}3$ kernels per stage followed by $2{\times}2$ max pooling (spatial resolution $32{\times}32 \rightarrow 16{\times}16 \rightarrow 8{\times}8$), then global average pooling and a linear classifier. The base width is $64$ channels in the first stage and $128$ in the second; there is no dropout. Inputs are per-channel standardized RGB images ($32{\times}32{\times}3$) with \emph{no data augmentation}. Optimization mirrors the HRM runs: AdamW with linear warmup (one epoch) followed by cosine decay with a $20\%$ floor, global-norm clipping $1.0$, label smoothing $\varepsilon=0.05$, learning rate $3{\times}10^{-4}$, weight decay $5{\times}10^{-4}$, batch size $128$, and $25$ training epochs. The final CNN has $273{,}344$ parameters. Typical epoch throughput is $\sim$391 steps at $\sim$31 it/s (about $12$ s/epoch).

\begin{figure}[tb]
  \centering
  \includegraphics[width=\linewidth]{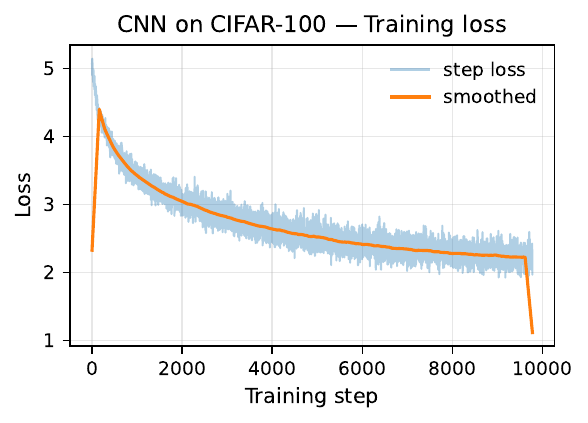}
  \caption{CNN on CIFAR-100: training loss per step (light) with a moving-average smoothed trace (dark).}
  \label{fig:cnn-cifar100-loss}
\end{figure}

\begin{figure}[tb]
  \centering
  \includegraphics[width=\linewidth]{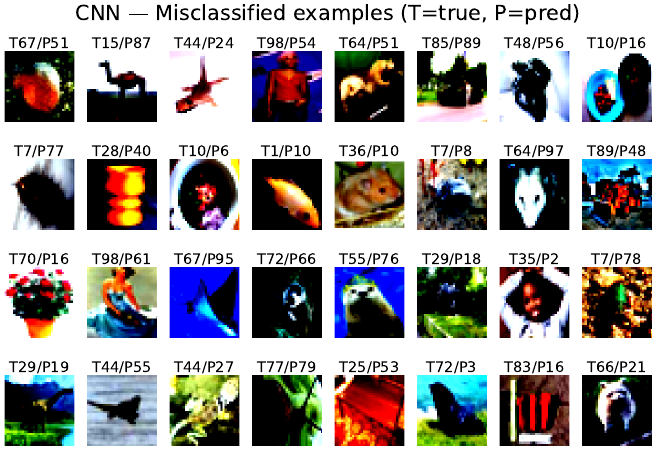}
  \caption{CNN on CIFAR-100: examples misclassified by the final model. Each tile shows ground-truth (\texttt{T}) and prediction (\texttt{P}).}
  \label{fig:cnn-cifar100-errors}
\end{figure}

Table~\ref{tab:cifar100-cnn-vs-hrm} aggregates the late-epoch metrics for the CNN alongside the HRM trained under the same \emph{raw} regime (no augmentation, identical optimizer family, label smoothing). The CNN improves steadily to \textbf{45.28\%} test accuracy by epoch~25 while maintaining moderate training accuracy ($\approx 0.50$), whereas HRM reaches near-saturated training accuracy ($\approx 0.92$) yet achieves only \textbf{29.70\%} on the test set. Combined with the per-epoch wall-time difference ($\sim$12~s vs.\ $\sim$5:52), this indicates that, for CIFAR-100 without augmentation, a simple convolutional architecture generalizes substantially better and is much more compute-efficient than HRM.

\begin{table}[t]
  \centering
  \begin{tabular}{@{}rcccccc@{}}
    \toprule
    & \multicolumn{3}{c}{\textbf{HRM}} & \multicolumn{3}{c}{\textbf{CNN}} \\
    \cmidrule(lr){2-4} \cmidrule(lr){5-7}
    \textbf{Epoch} & \textbf{Train loss} & \textbf{Train acc} & \textbf{Test acc} &
                      \textbf{Train loss} & \textbf{Train acc} & \textbf{Test acc} \\
    \midrule
    17 & 1.0823 & 0.8200 & ---    & 2.3797 & 0.4617 & 0.4228 \\
    18 & 1.0400 & 0.8350 & ---    & 2.3511 & 0.4690 & 0.4297 \\
    19 & 0.9941 & 0.8512 & ---    & 2.3265 & 0.4772 & 0.4352 \\
    20 & 0.9570 & 0.8629 & ---    & 2.3030 & 0.4824 & 0.4410 \\
    21 & 0.9277 & 0.8732 & ---    & 2.2834 & 0.4885 & 0.4465 \\
    22 & 0.8939 & 0.8859 & ---    & 2.2654 & 0.4920 & 0.4486 \\
    23 & 0.8659 & 0.8939 & ---    & 2.2504 & 0.4987 & 0.4585 \\
    24 & 0.8378 & 0.9065 & ---    & 2.2340 & 0.5039 & 0.4572 \\
    25 & 0.8104 & 0.9153 & \textbf{0.2970} & 2.2234 & 0.5045 & \textbf{0.4528} \\
    \bottomrule
  \end{tabular}
  \caption{CIFAR-100 (no augmentation). Late-epoch metrics for HRM and the CNN baseline under identical optimization settings and preprocessing. The CNN delivers higher test accuracy with far lower training accuracy (better generalization) and much faster epochs; HRM overfits (near-1.0 train accuracy) yet attains only $29.70\%$ test accuracy.}
  \label{tab:cifar100-cnn-vs-hrm}
\end{table}

\section{Discussion}

Across three image benchmarks under a deliberately 'raw' regime (no augmentation, identical optimizer family, label smoothing), the results show a consistent pattern: HRM trains stably but lacks the inductive bias needed to generalize on natural images, while small convolutional baselines are markedly more effective and far more compute‑efficient. On MNIST, HRM is competitive (98\% test accuracy), but on CIFAR‑10 it reaches only 65.0\% versus 77.2\% for a two‑stage CNN trained in needing only 12s/epoch (HRM needing 5:53/epoch). The gap widens on CIFAR‑100: HRM attains 29.7\% test after 25 epochs despite 91.5\% train accuracy (classic overfitting), whereas the same CNN achieves 45.3\% test with only 50.5\% train accuracy and the same streamlined training recipe, indicating substantially better inductive bias and regularization through locality/weight‑sharing. Qualitatively, the CNN’s training loss decreases smoothly throughout (and within‑epoch variance remains bounded), and its errors are distributed across visually confusable classes rather than collapsing on a few labels—both consistent with healthy optimization and moderate underfitting, not memorization (see the CNN loss trace and misclassified examples).

\bibliographystyle{plainnat}
\bibliography{references}

\end{document}